\documentclass[journal]{IEEEtran}
\usepackage{amsmath,amsfonts}
\usepackage{algorithmic}
\usepackage{algorithm}
\usepackage{array}
\usepackage[caption=false,font=normalsize,labelfont=sf,textfont=sf]{subfig}
\usepackage{textcomp}
\usepackage{stfloats}
\usepackage{url}
\usepackage{verbatim}
\usepackage{graphicx}
\usepackage{cite}
\usepackage{booktabs}
\usepackage{multirow}
\usepackage{float}
\usepackage{hyperref}
\begin{document}

\title{\fontsize{22pt}{25pt}\selectfont Document-Level Tabular Numerical Cross-Checking:\\
A Coarse-to-Fine Approach}


\author{Chaoxu Pang, Yixuan Cao, Ganbin Zhou, Hongwei Li, and Ping Luo

\thanks{
    Chaoxu Pang, Yixuan Cao, and Ping Luo are with Key Laboratory of Intelligent Information Processing, Institute of Computing Technology, Chinese Academy of Sciences. E-mail: \{pangchaoxu21b, caoyixuan, luop\}@ict.ac.cn
}
\thanks{
    Ganbin Zhou and Hongwei Li are with Beijing PAI Technology Ltd. E-mail: zhougb@paodingai.com, hw446.ict@gmail.com
}
\thanks{
    (Corresponding authors: Ping Luo and Yixuan Cao.)
}
}

\markboth{Journal of \LaTeX\ Class Files,~Vol.~14, No.~8, August~2021}%
{Shell \MakeLowercase{\textit{et al.}}: A Sample Article Using IEEEtran.cls for IEEE Journals}


\maketitle

\begin{abstract}
Numerical consistency across tables in disclosure documents is critical for ensuring accuracy, maintaining credibility, and avoiding reputational and economic risks. 
While prior research has primarily focused on instance-level fact checking, document-level verification remains underexplored despite its practical importance. 
Automated tabular numerical cross-checking presents two significant challenges: (C1) managing the combinatorial explosion of candidate instances at the document level and (C2) comprehending multi-faceted numerical semantics. 
Previous research typically depends on heuristic-based filtering or simplified context extraction, often struggling to balance performance and efficiency. 
Recently, large language models (LLMs) have demonstrated remarkable contextual understanding capabilities that helps address C2 at the instance level, yet they remain hampered by computational inefficiency (C1) and limited domain expertise. This paper introduces CoFiTCheck, a novel LLM-based coarse-to-fine framework that addresses these challenges through two sequential stages: embedding-based filtering and discriminative classification. 
The embedding-based filtering stage introduces an instructional parallel encoding method to efficiently represent all numerical mentions in a table with LLMs, as well as a decoupled InfoNCE objective to mitigate the isolated mention problem.
The discriminative classification stage employs a specialized LLM for fine-grained analysis of the remaining candidate pairs. This stage is further enhanced by our cross-table numerical alignment pretraining paradigm, which leverages weak supervision from cross-table numerical equality relationships to enrich task-specific priors without requiring manual annotation. Comprehensive evaluation across three types of real-world disclosure documents demonstrates that CoFiTCheck significantly outperforms previous methods while maintaining practical efficiency. Our approach represents a significant step forward in document-level verification, establishing an effective framework for tabular numerical cross-checking at scale.

\end{abstract}


\section{Introduction}\label{intro}


Fact checking refers to the process of comparing a claim with other sources of information to verify its accuracy. It has a wide range of applications, including fake news detection~\cite{shu2017fake} and claim verification in scientific publications~\cite{smeros2021sciclops}. As tables are important carriers of high-density information, fact checking in tabular contexts is particularly significant. However, existing table-based fact-checking studies~\cite{chen2020tabfact,Aly21Feverous,wang2024chain} primarily focus on instance-level verification of individual claims. In instance-level settings, each claim and its supporting table evidence are explicitly provided, allowing the system to focus on verifying a given claim-table pair. In contrast, document-level fact checking requires identifying and verifying relevant claim-table pairs from the entire document, where the number of candidate instances grows combinatorially. Document-level verification remains largely underexplored despite its significant real-world impact.
In this work, we address a challenging document-level fact-checking problem: \textit{verifying numerical consistency across tables in disclosure documents}. This task has broad applications in table-rich domains where numerical consistency is critical~\cite{hassan2017toward,sahitaj2025towards}.


In high-stakes domains such as finance, scientific research, government reporting, and corporate compliance, tables serve as the principal medium for presenting key quantitative indicators. Disclosure documents frequently contain extensive tabular data, where the same numerical fact may recur across different tables. We refer to these recurring numerical mentions as \textbf{semantically equivalent}. 
Figure \ref{fig:example_multi_faceted} illustrates this concept with screenshots of three tables from a corporate annual report. These tables present the indicators in a structured format with rows and columns, allowing readers to more easily comprehend and compare the underlying data. The numerical mentions highlighted with solid boxes across the three tables are semantically equivalent—they all represent the identical fact that \textit{the company's net assets at the end of fiscal year 2024 amounted to US\$49,120 million}. According to our statistics, over 20\% of numerical facts in disclosure documents are mentioned multiple times.

In practice, numerical inconsistencies among mentions of the same numerical fact can occur due to unintentional errors during document preparation. For instance, if any of the three highlighted numerical mentions in Figure \ref{fig:intro_example} were to deviate from the value ``49,120,'' an inconsistency would arise. Such errors can negatively impact public's perception and decision-making, potentially resulting in significant consequences. Several studies~\cite{Cao2018TowardsAN, wrong_numbers_risk_2019} have documented cases where numerical inaccuracies caused substantial reputational damage and economic losses across various sectors. As disclosure documents often form the backbone of transparency and regulatory compliance across industries, mechanisms for identifying and resolving numerical inconsistencies are essential for ensuring data integrity and public trust.


This situation underscores the pressing need for \textit{automated tabular numerical cross-checking systems}~\cite{guo2022survey}. The numerical cross-checking process can be decomposed into two sequential tasks: numerical semantic matching, which identifies all semantically equivalent numerical mention pairs within a document, and numerical comparison, which determines whether two such mentions are numerically equal. Pairs that are semantically equivalent but not numerically equal indicate potential inconsistencies. Since numerical comparison can typically be addressed with straightforward rules, this work concentrates on the more challenging task of \textbf{numerical semantic matching}. This task presents two primary challenges:

\begin{figure}[!t]
\centering
\subfloat[]{\includegraphics[width=0.45\textwidth]{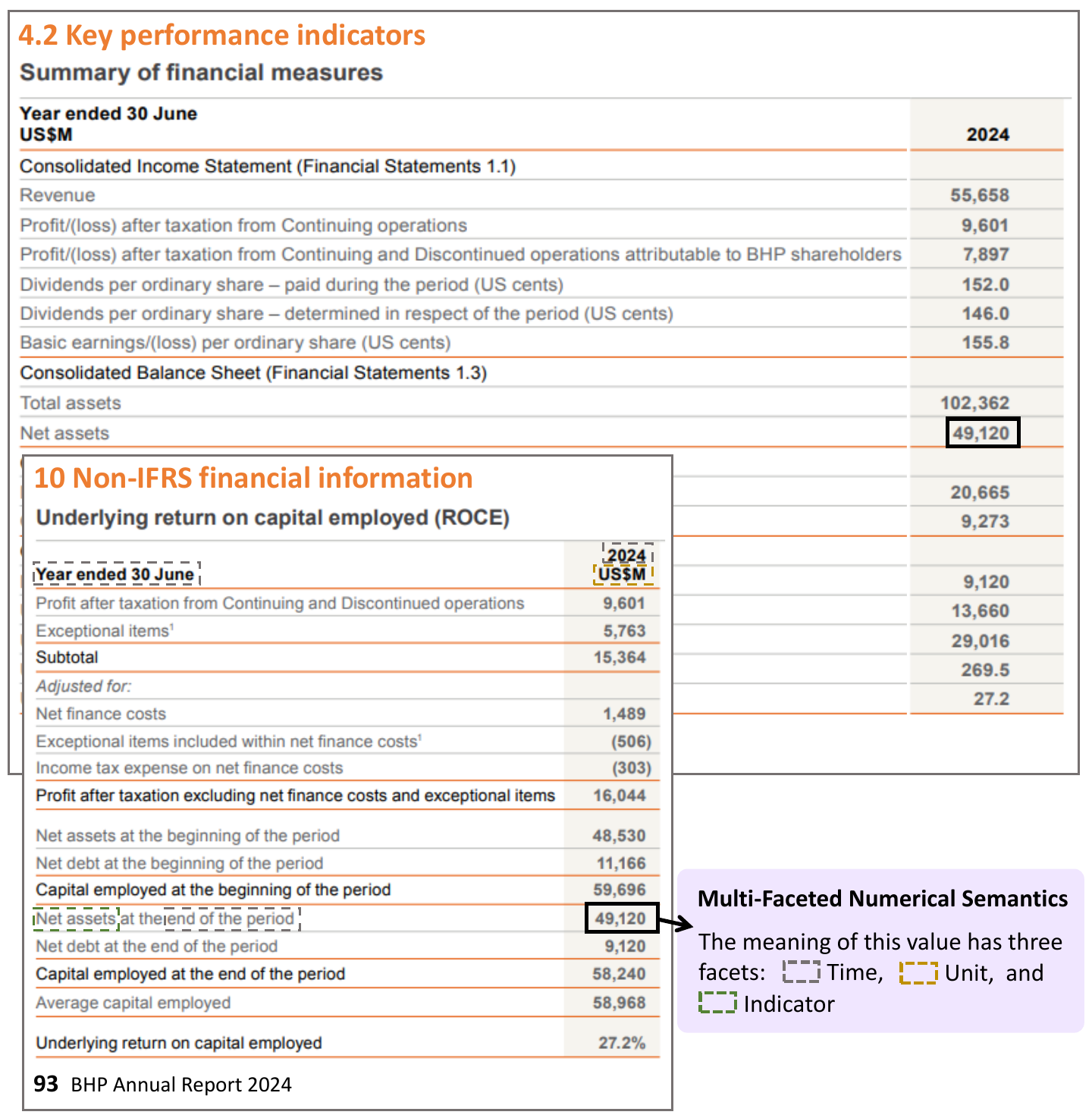}%
\label{fig:example_multi_faceted}}\\
\subfloat[]{\includegraphics[width=0.45\textwidth]{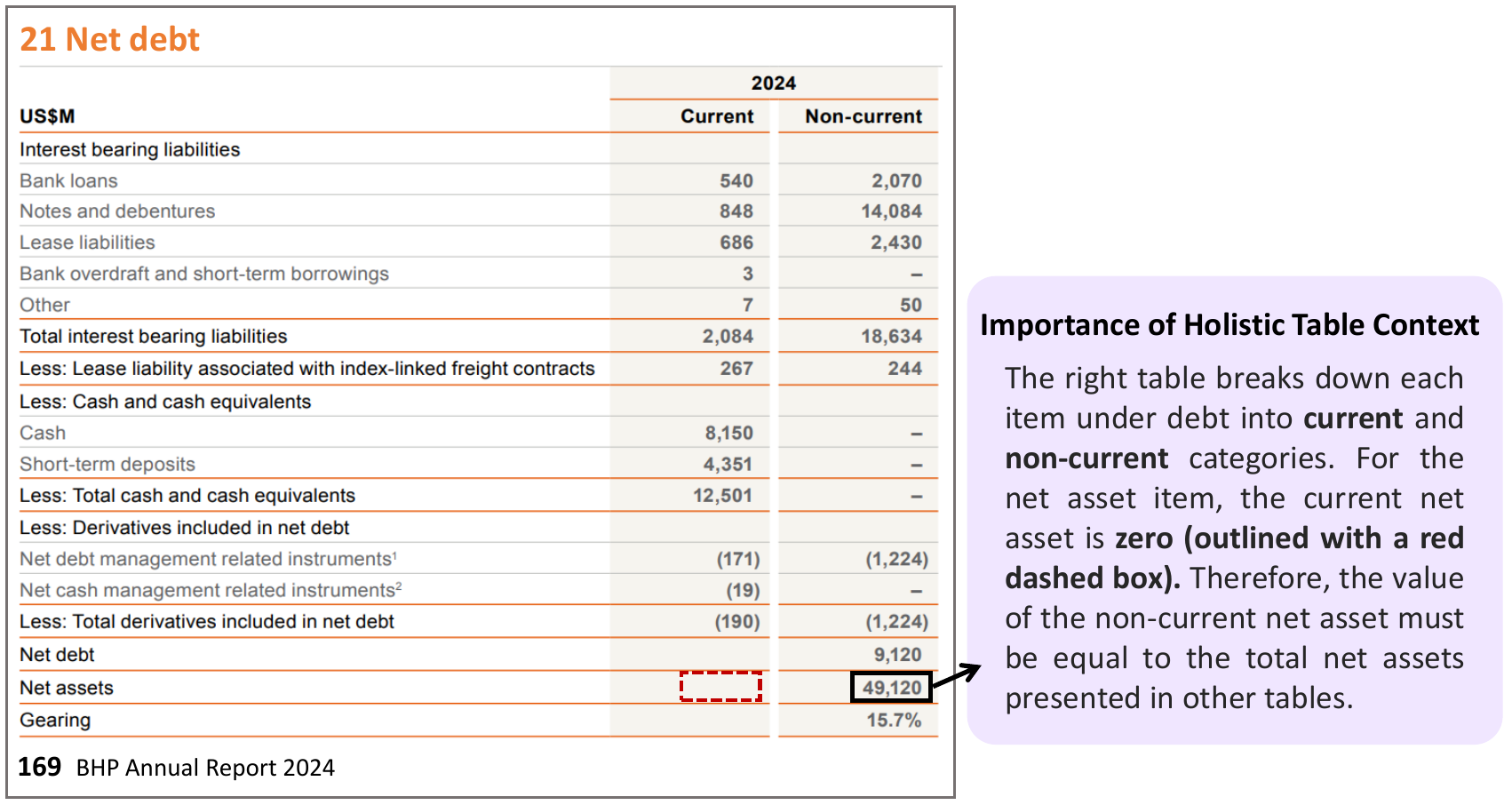}%
\label{fig:example_whole_table}}
\caption{Three tables from the BHP Annual Report 2024. Numerical mentions highlighted in black solid boxes are semantically equivalent across all three tables. This illustration also highlights: (a) the multi-faceted numerical semantics distributed within the table context; and (b) the importance of considering the holistic table context for precise numerical understanding.}
\label{fig:intro_example}
\end{figure}

\textbf{C1: Scalability for Massive Candidate Spaces.} While prior research has mainly investigated table-based fact checking~\cite{ye2023large} on instance-level datasets such as TabFact~\cite{chen2020tabfact}, there remains a significant gap in addressing document-level verification, where the candidate instance space within a single document grows combinatorially. In the context of numerical semantic matching, typical disclosure documents contain thousands of numerical mentions; because each mention must be compared with every other mention to assess semantic equivalence, a single document can yield millions of candidate mention pairs. This immense scale presents significant challenges for both computational efficiency and service timeliness, demanding methods that can effectively balance performance and efficiency at scale. Previous research~\cite{Li2020CrackingTP} has employed heuristic-based filtering techniques, such as grouping mentions along predefined attributes (e.g., time), which may improve efficiency but significantly limit the maximum achievable recall.

\textbf{C2: Multi-Faceted Numerical Semantics.} Each numerical mention encapsulates multiple semantic dimensions, such as temporal aspects and subject entities. The complete semantics extend beyond the surface-level values themselves, being distributed throughout the surrounding contexts---particularly within the table where the mention appears, as illustrated in Figure~\ref{fig:example_multi_faceted}. Previous research~\cite{Li2020CrackingTP} typically addresses the challenge by extracting simplified key contexts (e.g., row and column headers) with hand-crafted rules or shallow neural encoders. However, \textit{incorporating information from the complete table context is essential for comprehensive numerical semantic understanding.} For example, Figure~\ref{fig:example_whole_table} demonstrates that the information in one cell may influence the interpretation of another cell.


Recently, large language models (LLMs) such as GPT~\cite{gpt4} and Qwen~\cite{Yang2024Qwen2TR} have made remarkable progress in understanding context across both textual and semi-structured data~\cite{Liu2024LargeLM,jiang2023structgpt,wang2024chain}, creating exciting opportunities to tackle Challenge C2 at the instance level. However, the significant computational overhead and memory demands of LLMs~\cite{bommasani2021opportunities} introduce new efficiency bottlenecks for Challenge C1, particularly when processing real-world documents at scale and under service latency constraints. Moreover, general-purpose LLMs typically lack the specialized professional knowledge~\cite{fact_knowledge} required to accurately interpret numerical semantics within domain-specific contexts. For example, identifying the precise meaning of a numerical indicator may necessitate specific expertise in financial accounting, which is generally lacking in generic models.

To address these challenges, we introduce an efficient and high‑performing LLM‑based solution at the document level. We propose a novel \textbf{Co}arse-to-\textbf{Fi}ne \textbf{T}abular Numerical Cross-\textbf{Check}ing framework (CoFiTCheck), which operates through two sequential stages:

\textbf{Embedding-based Filtering.} We introduce an efficient embedding-based approach for filtering candidate numerical mention pairs. Each mention is encoded as a dense embedding, allowing us to prune potential pairs based on embedding similarity. To address the high computational cost of encoding large numbers of numerical mentions with LLMs~\cite{bommasani2021opportunities}, we introduce a \textit{contextualized instructional parallel encoding} strategy that jointly encodes all numerical mentions within a table in a single forward pass. For training, we propose a novel \textit{decoupled InfoNCE} objective tailored to the unique characteristics of numerical semantic matching, where isolated mentions (mentions without any semantic equivalent) are common and can distort the learning process.
Our decoupled approach explicitly accounts for both isolated and non-isolated mentions, enabling high-recall filtering while substantially reducing candidate pairs.

\textbf{Discriminative Classification.} We employ a larger, specialized LLM (ClsLLM) for fine-grained classification of remaining candidate mention pairs. To equip ClsLLM with domain-specific knowledge, we introduce \textit{Cross-table Numerical Alignment Pretraining (CNAP)}, a new pretraining paradigm that leverages cross-table numerical equality relationships as weak supervision signals, enabling the model to learn semantic equivalence patterns without manual annotation.

Comprehensive evaluation across three diverse types of real-world disclosure documents demonstrates the effectiveness and scalability of CoFiTCheck. Using a 7B parameter ClsLLM, our approach achieves approximately 90\% F1 score, surpassing previous methods by around 10 points. The framework exhibits remarkable efficiency, processing each document in just 40.8 seconds when deployed on four NVIDIA GeForce RTX 4090 GPUs. Notably, our CNAP approach delivers consistent performance gains without requiring manual annotations, highlighting its practical applicability. 
Overall, CoFiTCheck offers an effective solution for automated tabular numerical cross-checking in disclosure documents, delivering valuable insights for document-level fact checking in real-world applications.

\section{Related Work}

\subsection{Table-based Fact Checking}
Table-based fact checking (verification) has emerged as a critical research area in machine learning and natural language processing, serving as a primary defense against misinformation. 
Previous studies mainly focus on \textbf{statement-to-table checking}, which aims to determine whether natural language statements are entailed or refuted by tabular data. A significant line of research focuses on open-domain table fact checking. Datasets such as TabFact \cite{chen2020tabfact} and FEVEROUS \cite{Aly21Feverous} have catalyzed progress in this area, providing standardized benchmarks for developing and evaluating systems. Recently, Dater~\cite{ye2023large} proposed using large language models (LLMs) as versatile decomposers that break down complex statements into simpler components, combining this with a parsing-execution-filling strategy to decouple logic from numerical computation, achieving human-surpassing performance on the TabFact benchmark for the first time. These studies typically focus on the instance level, verifying statements against corresponding semi-structured tables from Wikipedia, and have achieved remarkable success in instance-level fact-checking.


Our work focuses on a distinct and challenging \textbf{document-level table-to-table checking} task: verifying the equivalence of numerical mentions in documents. This task presents two significant challenges: handling a large volume of candidate instances and understanding multi-faceted numerical semantics, as detailed in Section~\ref{intro}. The most recent work, AutoCheck \cite{Li2020CrackingTP}, addressed the first challenge by employing several grouping and deduplication rules to pre-filter candidate pairs. For the second challenge, the system first extracts key components of each numerical mention (such as row and column headers) and then encodes these components with a specialized cell embedding network. In real-world applications, AutoCheck demonstrated remarkable effectiveness, reducing auditing work hours by 52-68\%.
Despite its practical success, AutoCheck employed simplifications that limited its effectiveness. Specifically, it reduced complex table contexts to key parts and relied on heuristic rules to pre-filter candidate pairs. While these techniques enhanced system efficiency, they significantly compromised overall performance. To overcome these limitations, our current work introduces a coarse-to-fine approach that harnesses the power of LLMs, enabling us to preserve contextual richness without sacrificing computational efficiency.

\subsection{Large Language Models}

Large Language Models (LLMs)~\cite{gpt4o, o1, qwen2.5, v3} have emerged as a transformative force in recent years, demonstrating extensive world knowledge, strong contextual understanding, and sophisticated instruction-following capabilities. Our research intersects with two key sub-domains:

\textbf{LLMs for Representation Learning.} Recent research~\cite{luo2024large} has
revealed LLMs’ exceptional potential as backbone encoders over small models (e.g. BERT-based~\cite{devlin2019bert}) for dense retrieval tasks, largely due to their massive parameter counts and comprehensive pre-training regimes~\cite{ni2022large}. Several approaches~\cite{behnamghader2024llm2vec, zhuang2024promptreps} employ LLMs as unsupervised dense embedders—while computationally efficient, these methods often fail to fully leverage the models' inherent capabilities. More sophisticated strategies~\cite{ma2024fine, muennighoff2024generative} explicitly pre-train or fine-tune LLMs to optimize performance on retrieval tasks. For instance, Ma et al.~\cite{ma2024fine} fine-tuned LLaMA~\cite{touvron2023llama} models for multi-stage text retrieval, demonstrating significant improvements over smaller models and exhibiting impressive zero-shot capabilities.
While previous studies focus on encoding entire contexts or individual elements (queries, passages), our work focuses on the problem of simultaneously encoding multiple fine-grained facts (numerical mentions) within a shared context. We introduce a instructional parallel encoding approach that jointly represents all numerical mentions within a single table in one forward pass, substantially improving computational efficiency. Furthermore, we fine-tune LLMs using a decoupled InfoNCE objective specifically designed for numerical semantic matching tasks.

\textbf{LLMs for Table Understanding.} 
Recent studies have demonstrated that LLMs exhibit remarkable capabilities in understanding table semantics. 
Several comprehensive investigations provide \textbf{systematic evaluations} on table understanding abilities. 
Zhao et al.~\cite{zhao2023investigating} and Pang et al.~\cite{pang2024uncovering} highlight LLMs' effectiveness in information seeking from tabular data.
Akhtar et al.~\cite{akhtar2023exploring} evaluate LLMs' numerical reasoning capabilities across a hierarchical taxonomy of skills, finding that models such as GPT-3.5~\cite{gpt3.5} excel particularly in tabular natural language inference tasks, demonstrating their potential for numerical reasoning in structured contexts.
Beyond evaluations, numerous research efforts focus on \textbf{practical applications} that leverage and enhance LLMs' table understanding capabilities. 
Zhang et al.~\cite{zhang2024tablellama} show that LLMs significantly outperform smaller specialized models in table understanding tasks, with their TableLlama (fine-tuned on LLaMA 2~\cite{touvron2023llama}) achieving comparable or superior performance to state-of-the-art task-specific models across diverse table-based tasks. 
These findings collectively establish a strong foundation for our approach, which utilizes LLMs as powerful tools for interpreting and reasoning with tabular numerical semantics.
\section{Method}

\begin{figure*}[t]
\centering
\includegraphics[width=0.9\textwidth]{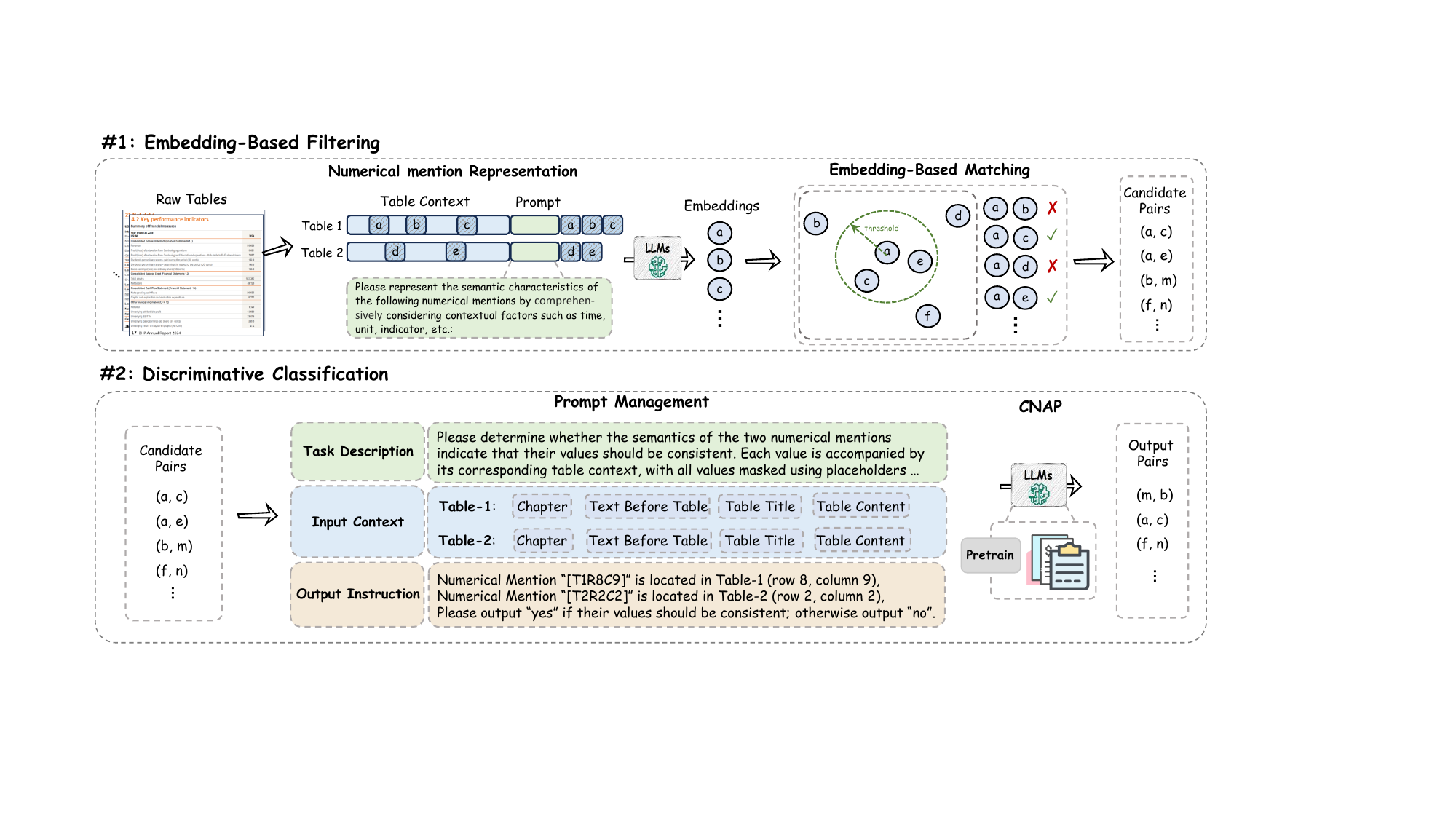}
\caption{The CoFiTCheck Framework. CoFiTCheck operates in two stages. First, for each input document, embedding-based filtering efficiently narrows down candidate numerical mention pairs from millions of possible combinations. Next, discriminative classification generates prompts for each candidate pair and performs fine-grained classification, leveraging prompt management and cross-table numerical alignment pretraining (CNAP). Lowercase letters are used to represent numerical mentions.}
\label{fig:method_overview}
\end{figure*}

\subsection{Overview}

Given a set of numerical mentions \( \mathcal{V} = \{v_k\}_{k=1}^{|\mathcal{V}|} \) and their associated table contexts \( \mathcal{C} = \{c_k\}_{k=1}^{|\mathcal{V}|} \) within a document, the numerical semantic matching task aims to identify semantically equivalent pairs of numerical mentions. 
The context \( c_k \) of a numerical mention \( v_k \) is a string that encompasses all relevant textual information required to interpret its semantics. Specifically, this context string \( c_k \) comprises the table containing \( v_k \) (typically linearized into markdown format~\cite{sui2024table}), the chapter title, surrounding text (limited to 500 characters) of the table, and the precise position of \( v_k \) within the tabular structure.


As illustrated in Figure~\ref{fig:method_overview}, CoFiTCheck addresses the numerical semantic matching task through two consecutive stages: \textit{Embedding-based Filtering} (Section~\ref{sec:emb_filter}) and \textit{Discriminative Classification} (Section~\ref{sec:disc_cls}). Each stage is powered by a specialized large language model - EmbLLM and ClsLLM respectively. 
Additionally, we introduce cross-table numerical alignment pretraining in Section~\ref{sec:pretraining} to further enhance ClsLLM's performance.


\subsection{Embedding-based Filtering}\label{sec:emb_filter}

When handling a large number of candidate numerical mention pairs within a single document (Challenge C1), it is crucial to efficiently reduce the search space before performing fine-grained classification (Section \ref{sec:disc_cls}). This is analogous to candidate item retrieval in recommendation systems~\cite{rec_1,rec_2,rec_3}, where an initial subset of potentially relevant items is retrieved before applying more computationally intensive reranking methods.

To prune candidate pairs, we propose an embedding-based approach that first encodes each numerical mention as a dense embedding. These embeddings capture compact semantic representations of the numerical mentions, enabling efficient retrieval of semantically equivalent pairs. 
Notably, we observe that a single table often contains multiple numerical mentions that share the same table context, differing only in their positions within the table. This motivates us to obtain embeddings for all mentions simultaneously rather than processing each one separately~\cite{ma2024fine,luo2024large}.
Furthermore, it is crucial to adhere to the instruction-tuning format that modern LLMs are predominantly trained on~\cite{ouyang2022training}. 
To this end, we propose a \textbf{Contextualized Instructional Parallel Encoding} (CIPE) strategy. 

Specifically, as illustrated in Figure~\ref{fig:method_overview}, we leverage an EmbLLM to encode all numerical mentions \{$v_{1}, ..., v_{n}$\} within a given table context $c$ in a single forward pass. We construct the input by concatenating the table context with a prompt $p_{\text{emb}}$ that instructs the LLM to encode the subsequent numerical mentions. All the numerical mentions within the table are then sequentially appended after the prompt. For each numerical mention $v_{j}$, we extract its representation $e_{j}$ by taking the last hidden state of its final token:
\begin{equation}
\label{emb_input}
[e_{1}, ..., e_{n}] = f_{\text{EmbLLM}}(c \oplus p_{\text{emb}} \oplus v_{1} \oplus ... \oplus v_{n}),
\end{equation}
where each component in Equation~\ref{emb_input} is first tokenized into a sequence of tokens before being fed into the model and $\oplus$ denotes sequence concatenation.
To prevent cross-contamination between numerical mentions after the prompt, we implement a specialized attention masking and positional encoding mechanism. Formally, for each numerical mention $v_{i}$ consisting of $N_i$ tokens $\{t_{i,1}, ..., t_{i,N_i}\}$, we modify the attention mask $M$ such that for $i \in [1,n], m \in [1,N_i]$:

\begin{equation}
M[t_{i,m}, t'] = 
\begin{cases}
1, & \text{if } t' \in \text{T}(c) \cup \text{T}(p_{\text{emb}}) \cup \{t_{i,1}, ..., t_{i,m}\} \\
0, & \text{otherwise}
\end{cases}
\end{equation}
where $\text{T}(\cdot)$ denotes the token set. This ensures that, after the prompt, tokens within numerical mentions can only attend to the table context $c$, the prompt $p_{\text{emb}}$, and preceding tokens within the same numerical mention. Additionally, we reset position indices for tokens of these numerical mentions to start after the end of the prompt $p_{\text{emb}}$, regardless of their absolute positions in the sequence:
\begin{equation}
\text{Position}(t_{i,m}) = |\text{T}(c) \cup \text{T}(p_{\text{emb}})| + m - 1. \\
\end{equation}
These adjustments preserve the contextual understanding while isolating the representations of individual numerical mentions.

After this step, we have embeddings of all numerical mentions \(\mathcal{E} = \{e_k\}_{k=1}^{|\mathcal{V}|}\). We then prune candidate pairs by retaining only those whose embeddings exhibit a similarity above a given threshold \( t \):
\begin{equation}
\mathcal{P}_{\text{cand}} = \{(i,j) \mid \cos(e_i, e_j) > t, (v_i,v_j) \in \mathcal{V}_i \times \mathcal{V}_j, i \neq j \}.
\end{equation}
To efficiently identify candidate pairs at scale, we leverage the HNSW algorithm~\cite{malkov2018efficient} implemented in the FAISS library~\cite{douze2024faiss} for approximate nearest neighbor searches across embeddings. This approach significantly reduces the computational complexity of constructing \(\mathcal{P}_{\text{cand}}\) from a naive \( O(|\mathcal{E}|^2) \) to approximately \( O(|\mathcal{E}|\log|\mathcal{E}|) \),  enabling efficient large-scale retrieval in practice.

To train the EmbLLM effectively, we propose a \textbf{decoupled InfoNCE} objective utilizing in-batch negatives~\cite{chen2020simple,in-batch-neg}. Each training batch consists of a collection of table contexts and their corresponding numerical mentions. For each mention $i$ in the batch, we define $\mathcal{P}(i)$ as the set of indices of mentions that are semantically equivalent to it, while treating the remaining mentions as negatives. 

Notably, numerical semantic matching differs from traditional retrieval tasks in two key aspects: (1) mentions serve as both queries and passages, and (2) most mentions are isolated without semantic equivalents. To address this, we propose a decoupled objective as follows. 
Let $\mathcal{N}{\text{n}}, \mathcal{N}{\text{i}}$ denote the set of non-isolated and isolated numerical mentions in a batch, respectively. Our training objective comprises two components:
\begin{align}
    \label{emb_loss}
    \mathcal{L}_{\text{n}} &= -\frac{1}{|\mathcal{N}_{\text{n}}|}\sum_{i=1}^{|\mathcal{N}_{\text{n}}|}\log\frac{\sum_{j\in \mathcal{P}(i)}{\exp(\mathrm{sim}(e_i,e_j)/\tau)}}{\sum_{k\in \mathcal{N}_{\text{n}}}{\exp(\mathrm{sim}(e_i,e_k)/\tau)}}, \\
    \mathcal{L}_{\text{i}} &= -\log\frac{\epsilon}{\epsilon+\sum_{\substack{(t,q)\in \mathcal{N}_{\text{i}} \times \mathcal{N}_{\text{i}} \\ t \neq q}}{\exp(\mathrm{sim}(e_t,e_q)/\tau)}},
\end{align}
where $\text{sim}(\cdot,\cdot)$ denotes the cosine similarity between two embeddings, $\tau$ is a temperature parameter, and $\epsilon$ is a small constant. The loss $\mathcal{L}_{\text{n}}$ encourages semantically equivalent mentions to have similar representations while pushing apart non-equivalent pairs, whereas $\mathcal{L}_{\text{i}}$ explicitly enforces dissimilarity between isolated numerical mentions. The final training objective is a weighted combination of these two losses: \( \mathcal{L} = \alpha_1 \mathcal{L}_{\text{n}} + \alpha_2 \mathcal{L}_{\text{i}} \).

\subsection{Discriminative Classification}\label{sec:disc_cls}
Following the coarse-grained embedding-based filtering stage, we perform fine-grained classification on each candidate pair to accurately determine their semantic equivalence. While embedding models effectively represent numerical mentions across the entire document space, they may not capture the nuanced semantic relationships between specific pairs. In contrast, discriminative classification examines one pair at a time, placing the contexts of both mentions together to form a query, allowing for more fine-grained comparative analysis of their semantic features. We conduct prompt management to facilitate LLMs in comprehending the task and producing the desired outputs. As illustrated in Figure~\ref{fig:method_overview}, the prompts are composed of the following three components:

\begin{itemize}
    \item \textbf{Task Description} $p_{\text{cls}}$: Provides explicit instructions for the task, including explanations of the input-output format and essential definitions to clarify specialized concepts relevant to the task.
    \item \textbf{Input Context}: Supplies the complete contextual information surrounding the two numerical mentions. Additionally, all numerical mentions in the table are masked with placeholders to prevent LLMs from relying on value equality as a shortcut for determining semantic equivalence.
    \item \textbf{Output Instruction} $p_{\text{out}}$: Specifies the locations (row and column) of the two target values in the table and instructs the LLM to make a binary decision on whether the two values are semantically equivalent.
\end{itemize}

Formally, for each pair $(i,j) \in \mathcal{P}_{\text{cand}}$, we prompt an LLM to perform fine-grained binary classification as follows:
\begin{equation}
r_{i,j} = f_{\text{ClsLLM}}(p_{\text{cls}} \oplus c_i \oplus c_j \oplus p_{\text{out}}(v_i, v_j)).
\end{equation}
We then parse the response $r_{i,j}$ to obtain the final prediction for each numerical mention pair. We train the ClsLLM using standard cross-entropy loss~\cite{radford2018improving}. 

\subsection{Cross-Table Numerical Alignment Pretraining}\label{sec:pretraining}

\begin{figure*}[t]
\centering
\includegraphics[width=0.85\textwidth]{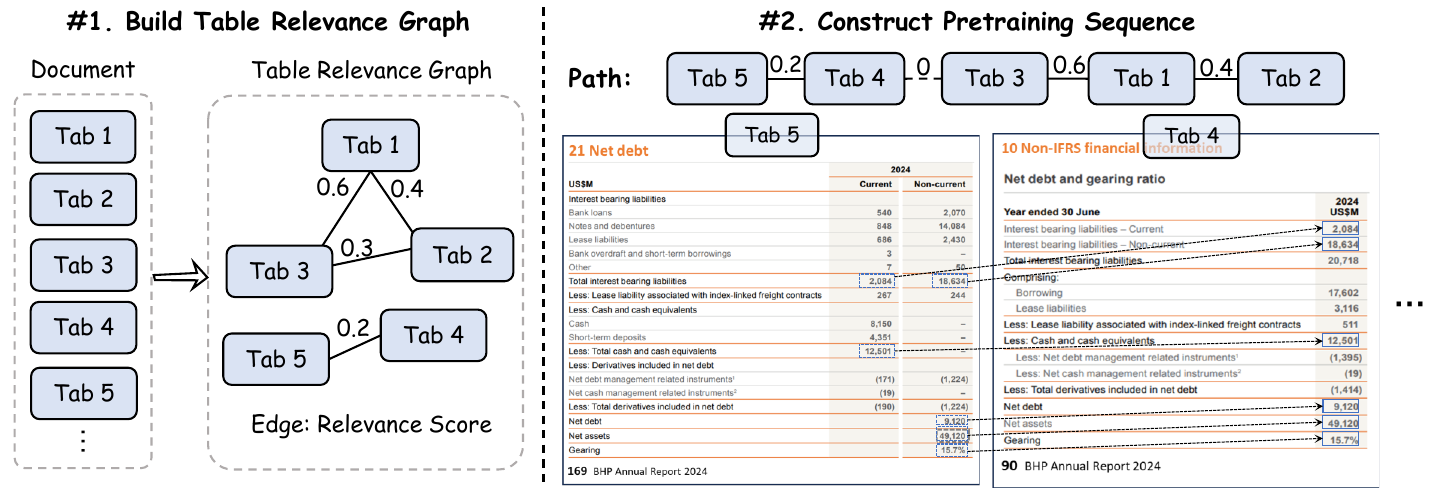}
 \caption{The workflow of Cross-Table Numerical Alignment Pretraining (CNAP). CNAP first builds a table relevance graph from a collection of tables in the document. Then, CNAP constructs a pretraining sequence by finding the maximum weight path that traverses all nodes exactly once.}
\label{fig:pretraining}
\end{figure*}

Though large language models (LLMs) exhibit broad world knowledge, recent studies~\cite{fact_knowledge} have shown that they still lack knowledge in professional domains. To enhance LLMs' understanding of semantically equivalent numerical mentions in professional documents, we propose a \textbf{C}ross-table \textbf{N}umerical \textbf{A}lignment \textbf{P}retraining (CNAP) approach. 

The key idea is that, rather than aligning numerical mentions with descriptions in natural languages, we aim to teach the model to identify patterns of semantically equivalent numerical pairs. Intuitively, we observe that \textit{the equality relationship between numerical mentions naturally provides weak supervision signals of semantic equivalence}. For example, in the bottom-right part of Figure~\ref{fig:pretraining}, when LLMs are pretrained to perform next-token prediction on the second (right) table, they are required to learn to identify and duplicate correct semantically equivalent values from the first (left) table. Therefore, we aim to reorder the tables in a document to construct pretraining sequences such that tables with more equal numerical mentions are positioned closer to each other. 

We formulate the problem as the maximum traveling salesman problem~\cite{flood1956traveling} which aims to find the maximum weight path that traverses all nodes in a graph exactly once.
Since exactly solving the problem is NP-hard, inspired by \cite{InContextPL}, we apply a greedy algorithm as an efficient approximation. 

As described in Algorithm~\ref{alg:cnap}, CNAP begins by representing tables in a document, denoted as $T=\{t_i\}_{i=1}^{|T|}$, as nodes in an undirected graph. The edges between nodes are weighted using relevance scores, which are defined as the Intersection over Union (IoU) of the numerical mention lists.
Mathematically, for two tables $t_i$ and $t_j$ with numerical mention lists $V_i$ and $V_j$, the relevance score can be expressed as:
\begin{equation}
R(t_i, t_j) = \frac{\text{equal}(V_i, V_j)}{|V_i| + |V_j|},
\end{equation}
where the \texttt{equal} function returns the number of equal numerical mentions.

The workflow of CNAP is depicted in Figure \ref{fig:pretraining}. CNAP traverses the graph by first selecting an unvisited table with the minimum degree as the starting node (Tab 5). Then the current path is iteratively extended by selecting the unvisited neighboring table with the highest weight (Tab 4). This process continues until reaching a node where all neighboring tables have been visited—a situation that could occur because the graph is not complete and only contains edges between tables sharing equal numerical mentions. In such cases, CNAP extends the graph with a zero-weight edge to a randomly selected unvisited table with the minimum degree (Tab 3) and resumes the process. The preference for selecting minimum-degree tables as starting points is because that they are most likely to have all their neighbors visited first, resulting in connections to irrelevant tables in the final path. Finally, the resulting traversal path is truncated to create fixed-sized input contexts appropriate for pretraining. We use the standard next token prediction loss~\cite{radford2018improving} for pretraining.

\begin{algorithm}[t]
\caption{CNAP}\label{alg:cnap}
\begin{algorithmic}[1]
\REQUIRE A document with tables $T=\{t_i\}_{i=1}^{|T|}$ and their numerical mention lists $\{V_i\}_{i=1}^{|T|}$
\ENSURE A pretraining dataset $\mathcal{P}_{pt}$

\STATE Initialize graph \(G = (T, E)\), where each table is a node, \(E\leftarrow \{\}\)
\FOR{each pair of tables \((t_i, t_j)\)}
    \STATE Compute the relevance score \(R(t_i, t_j) = \frac{\text{equal}(V_i, V_j)}{|V_i| + |V_j|}\)
    \IF{\(R(t_i, t_j) > 0\)}
        \STATE Add an edge \((t_i, t_j)\) to \(E\) with weight \(R(t_i, t_j)\)
    \ENDIF
\ENDFOR
\STATE Initialize path \(P \leftarrow []\)
\WHILE{\(|T| > 0\)}
    \STATE \(t_i \leftarrow \text{min\_deg}(T)\)
    \STATE \(P.\text{append}(t_i)\)
    \STATE \(T.\text{remove}(t_i)\)
    \WHILE{\(\text{Adj}(t_i) \cap T \neq \emptyset\)}
        \STATE \(t_j \leftarrow \arg \max_{t \in \text{Adj}(t_i) \cap T} \text{edge\_weight}(t_i, t)\)
        \STATE \(t_i \leftarrow t_j\)
        \STATE \(P.\text{append}(t_i)\)
        \STATE \(T.\text{remove}(t_i)\)
    \ENDWHILE
\ENDWHILE
\STATE Truncate \(P\) to create fixed-sized input contexts $\mathcal{P}_{pt}$
\RETURN \(\mathcal{P}_{pt}\)
\end{algorithmic}
\end{algorithm}

\section{experiments}

\subsection{Experimental Setups}
We first introduce basic experimental settings, including datasets, evaluation metrics, implementations, and baselines.

\subsubsection{Datasets}
We collect three sets of Chinese disclosure documents: \textit{IPO prospectuses}, \textit{auditor’s reports}, and \textit{annual reports}. These documents are widely used in financial disclosure and contain extensive tabular data, requiring a high degree of numerical consistency. \textit{IPO prospectuses} provide detailed information about a company’s financials and risks to ensure transparency and regulatory compliance during public offerings. \textit{Auditor’s reports} offer independent evaluations of financial statements, verifying their accuracy and enhancing stakeholder confidence. \textit{Annual reports} present a comprehensive summary of a company’s yearly performance, operations, and future outlook to inform shareholders and stakeholders. Audit reports cover the financials of a single year, and the financial sections of annual reports are mainly based on the data from the audit reports.
Each document is manually annotated using a pipeline similar to prior work~\cite{Li2020CrackingTP}. The statistics of these datasets are shown in Table~\ref{tab:data_stats}. Notably, the ratios of positive to negative pairs are highly imbalanced, particularly in auditor’s reports, which exhibit a pos-neg ratio of 1:73,362. This extreme imbalance poses significant challenges to the system’s performance and efficiency. 
After annotation, we split each dataset into training, validation, and test sets in an 8:1:1 ratio at the document level. Additionally, we crawled 11,635 annual reports from a stock exchange website~\footnote{https://www.szse.cn/disclosure/listed/notice/index.html} for pretraining purposes.

\begin{table}[ht]
    \centering
    \caption{Dataset Statistics. The abbreviation "pd." stands for "per document".}
    \label{tab:data_stats}
    \resizebox{\linewidth}{!}{%
    \begin{tabular}{lcccc}
        \toprule
        \textbf{Doc Type} & \textbf{\# Docs} & \textbf{\# Mentions pd.} & \textbf{\# Pos. Pairs pd.} & \textbf{Pos-Neg Ratio} \\
        \midrule
        IPO prospectuses & 270 & 4,546 & 973 & 1:24,015 \\
        Auditor’s reports & 762 & 8,825 & 1,130 & 1:73,362 \\
        Annual reports & 200 & 6,142 & 717 & 1:55,986 \\
        \bottomrule
    \end{tabular}%
    }
\end{table}

\subsubsection{Metrics}
The numerical semantic matching task aims to identify a set of semantically equivalent numerical pairs from a document. Due to the extreme imbalance in the ratios of positive to negative pairs, we adopt set-level precision, recall, and F1 as evaluation metrics, following prior work~\cite{Li2020CrackingTP}. Specifically, given a set of golden pairs \(\{g_1, ..., g_n\}\) and predicted pairs \(\{p_1, ..., p_n\}\) of \(n\) documents, we define the following metrics:

{\setlength{\jot}{8pt}
\begin{align}
\text{Precision (P)} & = \frac{\sum_{i=1}^{n}{|g_i \cap p_i|}}{\sum_{i=1}^{n}{|p_i|}}, \\
\text{Recall (R)}    & = \frac{\sum_{i=1}^{n}{|g_i \cap p_i|}}{\sum_{i=1}^{n}{|g_i|}}, \\
\text{F1}            & = \frac{2 \cdot \text{P} \cdot \text{R}}{\text{P} + \text{R}}.
\end{align}
}

\subsubsection{Implementations}
We select the Qwen2.5 series~\cite{qwen2.5} as our backbone due to its exceptional Chinese language understanding capabilities. 
For the embedding-based filtering stage, we utilize Qwen2.5-0.5B-Instruct as the backbone for EmbLLM. For the decoupled InfoNCE loss, we set the temperature $\tau$ to 0.15, $\alpha_1$ to 0.75, and $\alpha_2$ to 0.25. The model is trained for 3 epochs with a learning rate of $1\times10^{-5}$ and a batch size of 12 tables per GPU. During inference, we set the embedding similarity threshold to 0.5, which provides an effective balance between recall and efficiency (see Section~\ref{exp_emb} for detailed analysis).

For the discriminative classification stage, we adopt the 0.5B, 1.5B, 3B, and 7B instruct versions as backbones for ClsLLM, leveraging the increased model capacity to ensure more accurate classification. These models are trained for 2 epochs with a learning rate of $2\times10^{-5}$ and a batch size of 20 per GPU. For CNAP implementation on ClsLLM, we first pretrain the backbone for 2 epochs with a learning rate of $2\times10^{-5}$ and a batch size of 16 per GPU, followed by the same fine-tuning procedure.

All training procedures are conducted on a cluster of 24 H100 GPUs. We employ the Huggingface Transformers library~\cite{wolf-etal-2020-transformers} and DeepSpeed ZeRO~\cite{zero} for efficient distributed training, and utilize vLLM~\cite{vllm} for efficient inference. A cosine learning rate scheduler with linear warmup over the first 0.02 epochs is used. For both training and inference, the input length is set to 4096.

\subsubsection{Baselines}
For overall performance, we compare with the most recent work, \textbf{AutoCheck}~\cite{Li2020CrackingTP}. AutoCheck provides an end-to-end solution for the numerical semantic matching task. It first employs a cell embedding network to generate cell embeddings, followed by a cell pair classification step to determine whether each pair is semantically equivalent. To enhance efficiency, it applies heuristc-based filtering techniques, 
reducing the number of candidate pairs by a factor of four. We report the performance of AutoCheck as presented in the original paper~\cite{Li2020CrackingTP}. We use the same training and test splits, ensuring that the results are directly comparable.

Notably, AutoCheck is the only existing baseline specifically designed for this task. Recent studies have shown that for specific tasks, such as information extraction~\cite{guideline} and text classification~\cite{text-classification}, cutting-edge LLMs under zero/few-shot settings are comparable to or even surpass smaller expert models specifically trained on specialized tasks. For further comparison, we also evaluate two categories of state-of-the-art large language models without task-specific finetuning:
\begin{itemize}
    \item \textbf{General-purpose LLMs}, including GPT-4o-mini~\cite{gpt-4o-mini}, GPT-4o~\cite{gpt4o}, and DeepSeek-V3~\cite{v3}, which demonstrate broad world knowledge and remarkable table understanding capabilites.
    \item \textbf{Reasoning-specialized LLMs}, including OpenAI-o3-mini~\cite{o3-mini}, DeepSeek-R1~\cite{r1}, and OpenAI-o1~\cite{o1}, which are specifically optimized for advanced reasoning and excel in complex reasoning tasks. 
\end{itemize}

The baselines for embedding-based filtering and CNAP are described in detail in Sections~\ref{exp_emb} and~\ref{exp_cnap}, respectively.

\begin{table*}[t]
    \centering
    \caption{Overall Performance Comparison Across Different Document Types}
    \label{tab:sys_performance_comparison}
    \fontsize{6pt}{7.5pt}\selectfont
    \resizebox{16.5cm}{!}{%
    \begin{tabular}{lcccccccccc}
        \toprule
        \multirow{2}{*}{\textbf{Method}} & \multirow{2}{*}{\textbf{ClsLLM Size}} & \multicolumn{3}{c}{\textbf{Auditor's Reports}} & \multicolumn{3}{c}{\textbf{IPO Prospectuses}} & \multicolumn{3}{c}{\textbf{Annual Reports}} \\
        \cmidrule(lr){3-5} \cmidrule(lr){6-8} \cmidrule(lr){9-11}
        & & \textbf{P.} & \textbf{R.} & \textbf{F1.} & \textbf{P.} & \textbf{R.} & \textbf{F1.} & \textbf{P.} & \textbf{R.} & \textbf{F1.} \\
        \midrule
        AutoCheck~\cite{Li2020CrackingTP} & - & 74.8 & 76.0 & 75.4 & 84.6 & 78.3 & 81.3 & - & - & - \\
        CoFiTCheck & 0.5B & 83.2 & 84.4 & 83.8 & 89.3 & 84.3 & 86.7 & 86.9 & 90.7 & 88.7 \\
        CoFiTCheck & 1.5B & 84.7 & 85.0 & 84.8 & 89.1 & 84.9 & 87.0 & 88.4 & 90.6 & 89.5 \\
        CoFiTCheck & 3B & 85.7 & 86.7 & 86.2 & 91.8 & 87.0 & 89.4 & 89.5 & 91.4 & 90.4 \\
        CoFiTCheck & 7B & 87.0 & \textbf{86.9} & 87.0 & \textbf{92.8} & \textbf{87.9} & \textbf{90.3} & 90.0 & 91.6 & 90.8 \\
        \cmidrule(lr){1-11}
        \multicolumn{11}{l}{\textit{With additional pretraining documents}} \\
        CoFiTCheck w. CNAP & 0.5B & 83.9 & 84.5 & 84.2 & 88.9 & 83.9 & 86.3 & 87.4 & 90.6 & 89.0 \\
        CoFiTCheck w. CNAP & 1.5B & 85.2 & 85.8 & 85.5 & 90.4 & 86.1 & 88.2 & 89.0 & 90.7 & 89.8 \\
        CoFiTCheck w. CNAP & 3B & 86.8 & 86.6 & 86.7 & 92.0 & 86.8 & 89.3 & 89.7 & 91.9 & 90.7 \\
        CoFiTCheck w. CNAP & 7B & \textbf{87.5} & 86.8 & \textbf{87.1} & 92.7 & 87.7 & 90.1 & \textbf{91.0} & \textbf{91.9} & \textbf{91.5} \\
        \bottomrule
    \end{tabular}%
    }
\end{table*}

\begin{figure*}[t]
\centering
\includegraphics[width=0.92\textwidth]{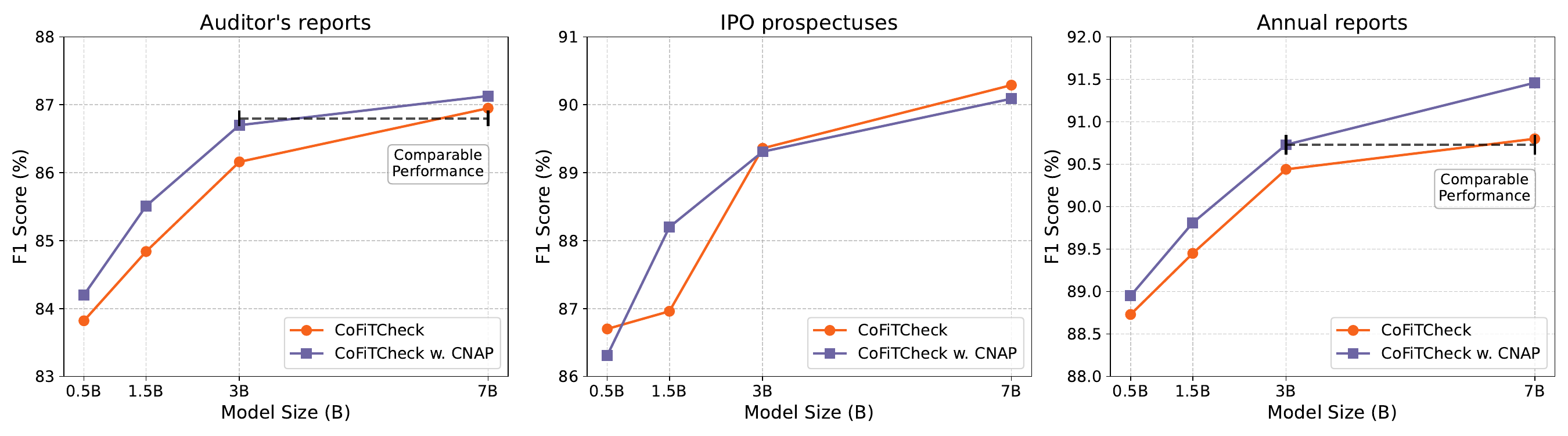}
\caption{F1 scores of CoFiTCheck across three document types with varying ClsLLM sizes. CoFiTCheck w. CNAP generally outperforms CoFiTCheck, with performance consistently improving as ClsLLM size increases from 0.5B to 7B parameters.}
\label{fig:performance}
\end{figure*}

\subsection{Overall Performance Comparison}\label{sec:sys_results}

In this section, we present the overall performance of the numerical semantic matching task on three types of documents.
As shown in Table~\ref{tab:sys_performance_comparison} and Figure~\ref{fig:performance}, we can observe several key findings:

\begin{itemize}
    \item \textbf{Superior performance of CoFiTCheck:} Our proposed CoFiTCheck  significantly outperforms AutoCheck across all document types. With the 0.5B ClsLLM backbone, CoFiTCheck achieves an F1 score of 83.8\% and 86.7\% on auditor's reports and IPO prospectuses, surpassing AutoCheck by 8.4 and 5.4 points, respectively.
    Increasing the size of ClsLLM significantly improves performance. CoFiTCheck with 7B ClsLLM achieves highest performance, reaching 87.0\%, 90.3\%, and 90.8\% F1 scores on auditor's reports, IPO prospectuses, and annual reports, respectively. This represents an improvement of 11.6 points over AutoCheck on auditor's reports and 9.0 points on IPO prospectuses.
    
    \item \textbf{Effectiveness of CNAP:} Our proposed Cross-Table Numerical Alignment Pretraining (CNAP) method demonstrates effectiveness in further boosting overall  performance without manual annotations. For example, when applied to the 1.5B ClsLLM, CNAP improves the F1 scores by 0.7, 1.2, and 0.3 points across the three document types compared to the standard CoFiTCheck with the same backbone size. \textit{CNAP enables smaller ClsLLM models to achieve performance comparable to their larger counterparts without CNAP.} As shown in Figure \ref{fig:performance}, the 3B model with CNAP achieves 86.7\% and 90.7\% F1 scores on auditor's reports and annual reports respectively, which is comparable to the 7B model without CNAP (87.0\% and 90.8\%). 
    Notably, CoFiTCheck with CNAP using the 7B ClsLLM backbone achieves the best overall performance on auditor's reports and annual reports. CNAP demonstrates more substantial performance gains on annual reports and auditor's reports compared to IPO prospectuses. This likely stems from our pretraining corpus composition, which predominantly consists of 11,635 annual reports\footnote{Compared to IPO prospectuses, annual reports and auditor's reports share more similar tabular structures and financial indicators.}. We consider expanding our pretraining to incorporate a more diverse range of document types as an important direction for future work.
\end{itemize}

These results validate the effectiveness of the CoFiTCheck framework and the proposed pretraining method CNAP for numerical semantic matching tasks over disclosure documents, particularly when combined with larger model capacities.

\subsection{Performance Comparison with SOTA LLMs}\label{exp_cls_model}

In this section, we compare our ClsLLM with state-of-the-art LLMs on the discriminative classification task.
We randomly select 1k samples from the test set of discriminative classification as the test bench\footnote{We limit the test set size to reduce API calling costs when evaluating commercial LLMs.}. For all models, we use greedy decoding (temperature = 0) with zero-shot prompting. The results are shown in Table \ref{tab:clsllm_performance}.

\begin{table}[htbp]
\centering
\caption{Performance Comparison of Various LLMs on Discriminative Classification}
\label{tab:clsllm_performance}
\resizebox{7.5cm}{!}{%
\begin{tabular}{lcccc}
\toprule
\textbf{Model} & \textbf{Acc.} & \textbf{P.} & \textbf{R.} & \textbf{F1.} \\
\midrule
\multicolumn{5}{l}{\textit{General-purpose LLMs}} \\
GPT-4o-mini~\cite{gpt-4o-mini} & 61.1 & 87.5 & 1.8 & 3.5 \\
GPT-4o~\cite{gpt4o}     & 72.7 & 67.9 & 58.5 & 62.9 \\
DeepSeek-V3~\cite{v3} & 78.4 &	70.1 &	79.0 &	74.3 \\
\midrule
\multicolumn{5}{l}{\textit{Reasoning-specialized LLMs}} \\
OpenAI-o3-mini~\cite{o3-mini}     & 74.6 & 66.7 & 71.4 & 69.0 \\
DeepSeek-R1~\cite{r1}       & 79.8 & 72.8 & 78.0 & 75.3 \\
OpenAI-o1~\cite{o1}         & 82.0 & 76.8 & 78.0 & 77.4 \\
\midrule
\multicolumn{5}{l}{\textit{Ours}} \\
ClsLLM-0.5B & 89.5 & 84.0 & 90.6 & 87.2 \\
ClsLLM-0.5B w. CNAP & 90.3 & 84.8 & 91.9 & 88.2 \\
ClsLLM-7B & 92.9 & 89.1 & \textbf{93.4} & 91.2 \\
ClsLLM-7B w. CNAP  & \textbf{93.1} & \textbf{90.0} & 92.9 & \textbf{91.4} \\
\bottomrule
\end{tabular}
}
\end{table}

The experimental results reveal several important findings:

\begin{itemize}
    \item \textbf{SOTA LLMs show promising performance:} SOTA LLMs demonstrate strong capabilities in the discriminative classification task without specific fine-tuning, with OpenAI-o1 achieving an F1 score of 77.4\%. This indicates that recent advancements in LLMs have equipped these models with great numerical understanding abilities in tables. Notably, reasoning-specialized models consistently outperform general-purpose counterparts from the same provider. This performance gap likely stems from the nature of the discriminative classification task, which requires analyzing and comparing numerical semantics in tables—a process inherently demanding reasoning capabilities.
    
    \item \textbf{Task-specific fine-tuning remains crucial:} The 0.5B ClsLLM significantly outperforms the best reasoning-specialized model, OpenAI-o1. The advantage becomes even more pronounced with ClsLLM-7B w. CNAP, which achieves an F1 score improvement of 14 points. Examining the false positive rate (i.e., $1-\text{precision}$) further highlights this gap: OpenAI-o1 exhibits a false positive rate of 22.6\%, whereas ClsLLM-7B w. CNAP reduces this to just 8.6\%, representing an almost threefold decrease. This considerable performance gap underscores that discriminative classification demands specialized knowledge and domain expertise that current LLMs lack, highlighting the importance of task-specific fine-tuning even in the era of powerful foundation models.
    
\end{itemize}

\subsection{Overall Efficiency Comparison}\label{exp_efficiency}

In this section, we evaluate the efficiency of CoFiTCheck using 126 test documents on 4 NVIDIA GeForce RTX 4090 GPUs. We deploy the EmbLLM and ClsLLM sequentially as 4 distributed workers across these GPUs, reporting the averaged per-document processing time for each stage. Our ablation studies examine: (1) \textbf{Removing Parallel Encoding}, removing the parallel encoding strategy of CIPE, which forces encoding one numerical mention per forward pass;
(2) \textbf{Heuristic-based Filtering}, replacing stage 1 with heuristic-based filtering from AutoCheck~\cite{Li2020CrackingTP}; and (3) \textbf{Removing Stage 1}, removing embedding-based filtering entirely, which processes all candidate pairs in stage 2. For the latter two computationally intensive scenarios, we estimate\footnote{Since these experiments are too time-consuming, we don't run the complete experiments and report the estimated time according to the average processing time.} runtimes based on average processing times. 

\begin{table}[ht]
\centering
\caption{Runtime comparison of CoFiTCheck system across different ClsLLM sizes, showing average processing time per document.}
\label{tab:runtime_comparison}
\resizebox{8.2cm}{!}{%
\begin{tabular}{lrrr}
\toprule
\textbf{Method} & \textbf{Stage 1 (sec)} & \textbf{Stage 2 (sec)} & \textbf{Total (sec)} \\
\midrule
AutoCheck~\cite{Li2020CrackingTP} & - & - & 166.6 \\
CoFiTCheck (0.5B) & 12.4 & 3.3 & 15.7 \\
CoFiTCheck (7B) & 12.4 & 28.4 & 40.8 \\
\cmidrule(lr){1-4}
\multicolumn{4}{l}{\textit{Removing Parallel Encoding}} \\
CoFiTCheck (0.5B) & 309.9 & 3.3 & 313.2 \\
CoFiTCheck (7B) & 309.9 & 28.4 & 338.3 \\
\cmidrule(lr){1-4}
\multicolumn{4}{l}{\textit{Heuristic-based Filtering}} \\
CoFiTCheck (0.5B) & 0 & 6,623.3 & 6,623.3 \\
CoFiTCheck (7B) & 0 & 57,000.3 & 57,000.3 \\
\cmidrule(lr){1-4}
\multicolumn{4}{l}{\textit{Removing Stage 1}} \\
CoFiTCheck (0.5B) & 0 & 129,399.1 & 129,399.1 \\
CoFiTCheck (7B) & 0 & 1,113,616.5 & 1,113,616.5 \\
\bottomrule
\end{tabular}
}
\end{table}

As presented in Table \ref{tab:runtime_comparison}, CoFiTCheck demonstrates remarkable efficiency. It processes a document in just 15.7 seconds with the 0.5B ClsLLM and 40.8 seconds with the 7B ClsLLM. Considering that manual verification typically requires tens of hours of expert review per document~\cite{Li2020CrackingTP}, CoFiTCheck's processing speed is well-suited for practical deployment in real-world scenarios.

Our ablation study reveals several key efficiency insights:

\begin{itemize}
    \item \textbf{Superior efficiency of Parallel Encoding:} When the parallel encoding strategy is removed, the processing time for stage 1 increases dramatically from 12.4 seconds to 309.9 seconds—a 25$\times$ slowdown—highlighting the effectiveness of our parallel encoding approach. A similar acceleration is observed during training: with parallel encoding, the training process takes approximately 1 day, whereas without it, training would require about 25 days.
    
    \item \textbf{Necessity of embedding-based filtering:} When stage 1 is removed entirely, the processing time increases to approximately 1.5 days for the 0.5B model and 12.9 days for the 7B model. Besides, CoFiTCheck with embedding-based filtering is approximately 420× faster than using heuristic-based filtering with the 0.5B model and 1,400× faster with the 7B model.
\end{itemize}

These improvements collectively address Challenge C1 for tabular numerical cross-checking, making CoFiTCheck practical for real-world applications.

\subsection{Analysis of Embedding-Based Filtering}\label{exp_emb}

Embedding-based filtering plays a crucial role in enhancing system efficiency by pruning candidate pairs. However, this process may inadvertently exclude true positive pairs, thereby affecting the overall system recall. This section analyzes the trade-off between computational efficiency and recall across various embedding similarity thresholds.

We conduct the following comparison experiments to validate our design choices for EmbLLM: (1) \textbf{standard InfoNCE}. We compare decoupled InfoNCE objective with the standard InfoNCE objective~\cite{izacard2021unsupervised}, which is formulated as:
\begin{equation}
    \mathcal{L}_{\text{standard}} = -\frac{1}{N}\sum_{i=1}^{N}\log\frac{\sum_{j\in \mathcal{P}(i)}{\exp(\text{sim}(e_i,e_j)/\tau)}}{\sum_{k=1}^{N}{\exp(\text{sim}(e_i,e_k)/\tau)}},
\end{equation}
where $N$ is the number of numerical mentions in a batch and $\mathcal{P}(i)$ represents the set of positive numerical mentions for the $i$-th mention. We treat each mention as a positive mention of itself, ensuring that every mention has at least one positive mention. Standard InfoNCE treats all mentions equally and double-counts isolated mention pairs, which may disproportionately focus on distinguishing isolated mentions rather than bringing positive pairs closer.
(2) \textbf{Extractive parallel encoding (EPE)}. We compare our contextualized instructional parallel encoding (CIPE) strategy with the extractive parallel encoding (EPE) strategy~\cite{muennighoff2024generative}, which directly encodes the table context and uses the embedding of the last token of each numerical mention as its numerical representation, without adding additional prompt and mention tokens in our method. (3) \textbf{Decoupled InfoNCE w/o. $\mathcal{L}_i$}. We remove the loss term $\mathcal{L}_i$ in Equation \ref{emb_loss}.

\begin{figure*}[!t]
\centering
\subfloat[]{\includegraphics[width=0.335\textwidth]{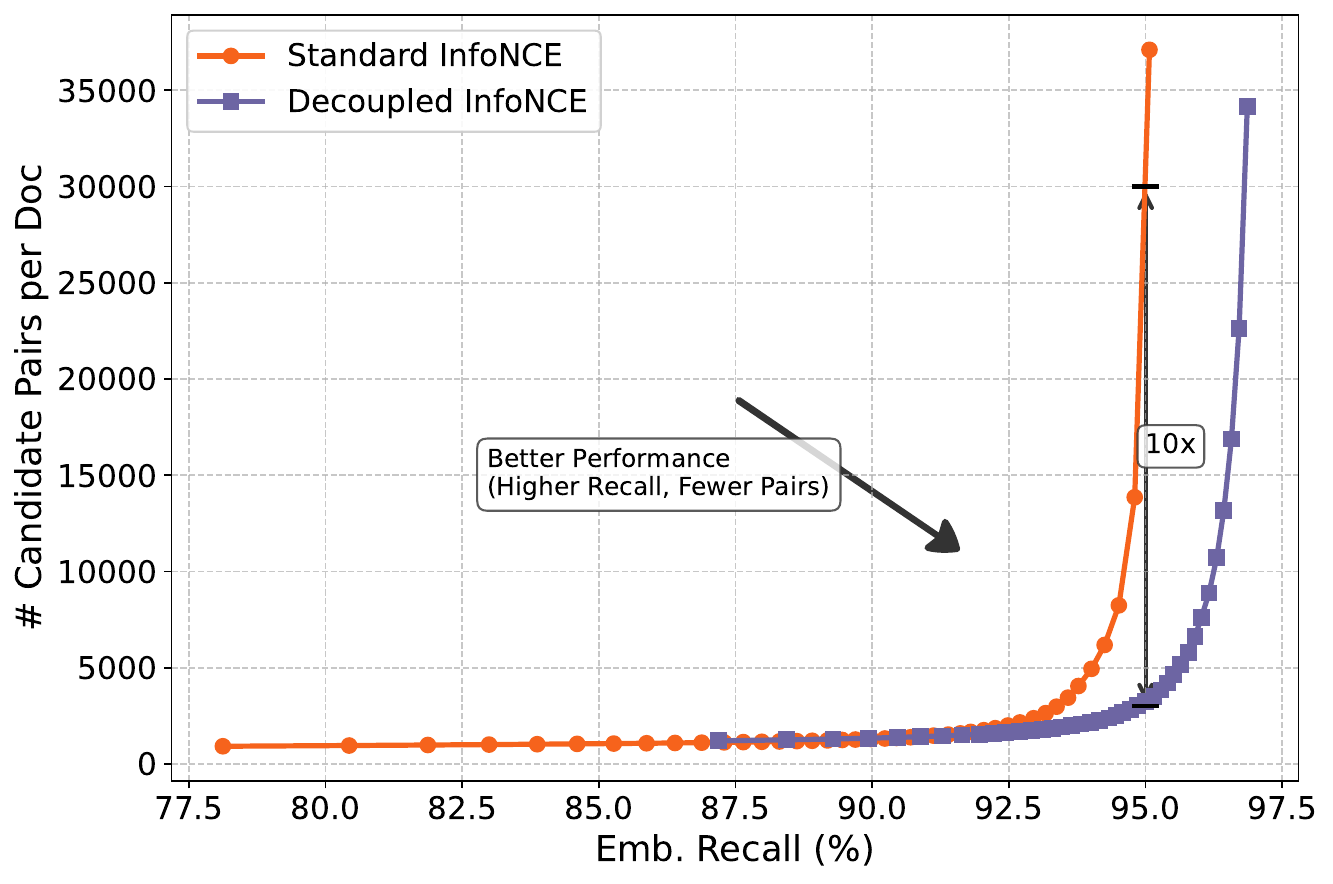}%
\label{fig:emb_comp}}
\subfloat[]{\includegraphics[width=0.32\textwidth]{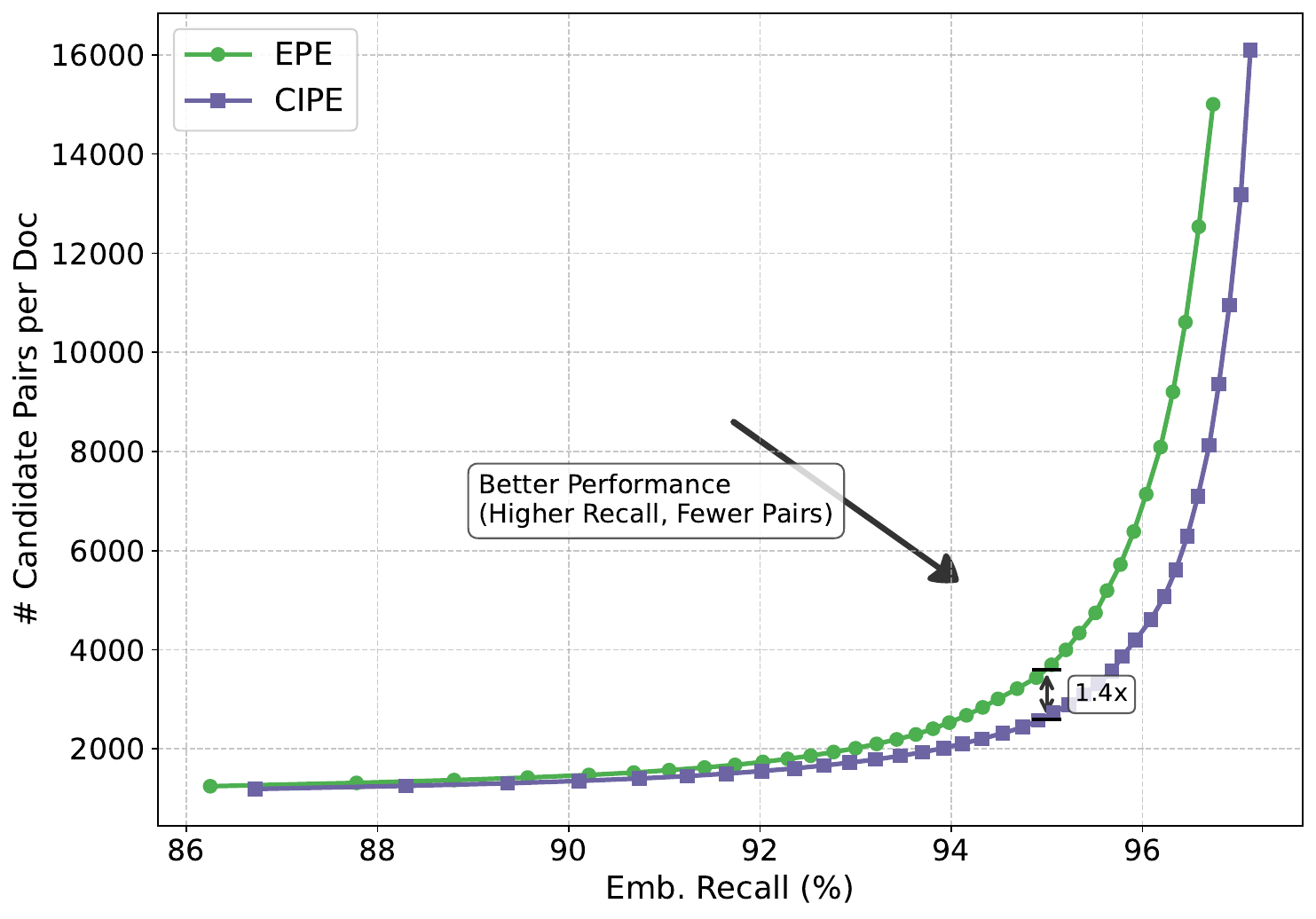}%
\label{fig:emb_comp_extractive}}
\subfloat[]{\includegraphics[width=0.32\textwidth]{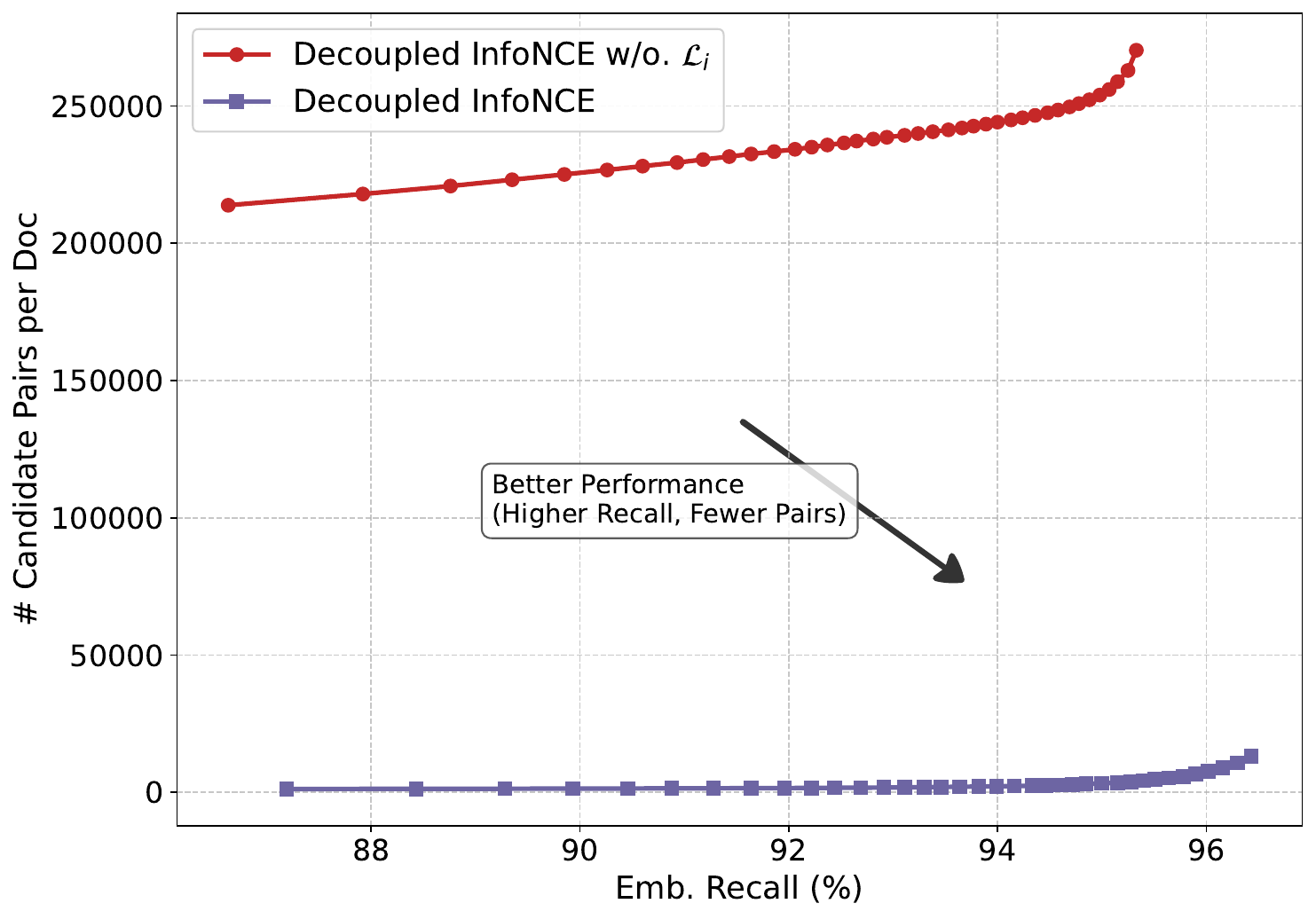}%
\label{fig:emb_comp_abl_loss}}
\caption{Comparison of EmbLLM performance under different settings: (a) comparison between our decoupled InfoNCE objective and the standard InfoNCE objective; (b) comparison between our CIPE strategy and the EPE strategy; and (c) ablation study of the decoupled InfoNCE objective with and without the loss term $\mathcal{L}_i$. In all plots, the x-axis represents recall (higher is better), and the y-axis indicates the number of candidate pairs per document (lower is better).}
\label{fig_sim}
\end{figure*}

Using 126 test set documents as benchmark, we vary the embedding similarity threshold from 0.1 to 0.9, measuring both the remaining candidate pairs per document and the recall. As illustrated in Figure~\ref{fig_sim}, more aggressive filtering (higher threshold) reduces candidate pairs but potentially decreases recall by filtering out true positives. A practical balance between recall and efficiency lies in the lower-right region, indicating relatively high recall while keeping the number of remaining candidate pairs low.

Our analysis yields the following key findings:
\begin{itemize}
    \item \textbf{Superior performance of decoupled InfoNCE:} The decoupled InfoNCE objective consistently outperforms the standard InfoNCE objective (Figure~\ref{fig:emb_comp}), producing fewer candidate pairs to achieve equivalent recall levels. This advantage is especially pronounced at higher recall settings, which are critical for practical applications.
    At a 95\% recall level, our decoupled objective outputs only one-tenth of the candidate pairs compared to the standard objective. This substantial reduction would deliver a nearly 10$\times$ speedup in both training and inference for the downstream ClsLLM module, significantly enhancing overall efficiency. 
    Additionally, removing the loss term $\mathcal{L}_i$ (Figure~\ref{fig:emb_comp_abl_loss}) results in a marked performance decline, underscoring the critical role of $\mathcal{L}_i$ in effectively pushing apart isolated mentions.
    \item \textbf{CIPE outperforms EPE:} The CIPE strategy consistently surpasses EPE across all recall levels (Figure~\ref{fig:emb_comp_extractive}). At a 95\% recall level, our CIPE strategy outputs only two-thirds of the candidate pairs compared to the EPE strategy. This improvement is likely because our proposed CIPE strategy employs an instruction format that is more closely aligned with the training paradigms of LLMs, such as instruction tuning~\cite{ouyang2022training}.
\end{itemize}

\subsection{Analysis of CNAP}\label{exp_cnap}

As described in Sections \ref{sec:sys_results}, our proposed Cross-Table Numerical Alignment Pretraining (CNAP) method consistently boosts overall performance 
without requiring manual annotations. In this section, we investigate the contributions of two key components to CNAP's effectiveness: (1) the additional pretraining process and (2) the advanced pretraining sequence construction strategy. 
Specifically, we compare CNAP with a \textbf{R}eading \textbf{O}rder-aware \textbf{P}re\textbf{T}raining (ROPT) strategy, which is widely adopted as a robust recipe for pretraining generative language models~\cite{llama3,qwen2.5,pt_1}. For each document, ROPT constructs pretraining sequences using the same tables as CNAP but traverses them following the document's reading order. The training recipe for ClsLLM remains identical between ROPT and CNAP. We employ the 1.5B backbone to compare these methods and report overall F1 scores.

\begin{figure}[ht]
\centering
\includegraphics[width=0.38\textwidth]{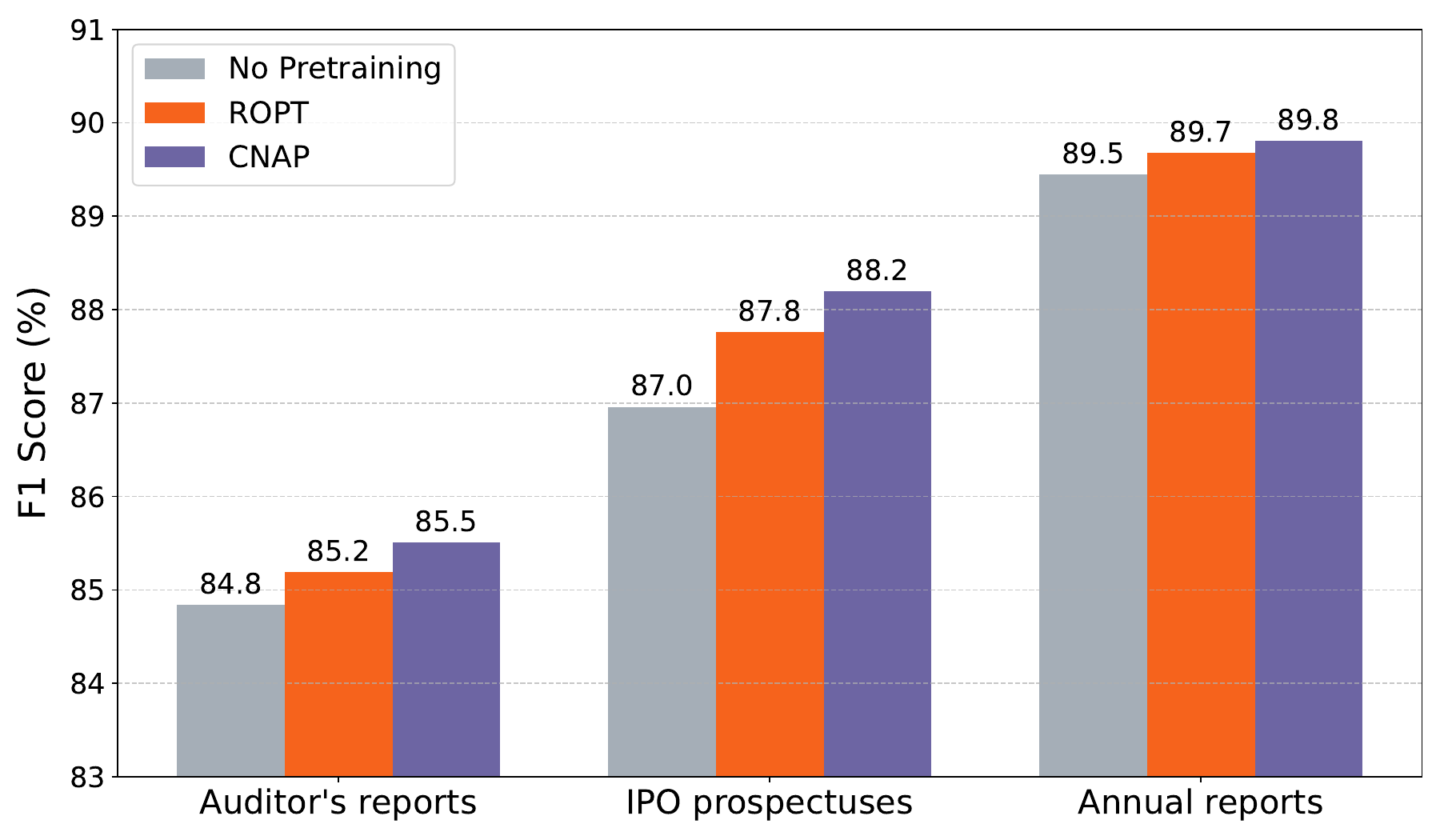}
\caption{Overall F1 scores of various pretraining strategies (1.5B parameters) evaluated across three document types.}
\label{fig:exp_cnap}
\end{figure}

As shown in Figure \ref{fig:exp_cnap}, ROPT improves over the baseline without pretraining by 0.4, 0.8, and 0.2 F1 points on auditor's reports, IPO prospectuses, and annual reports, respectively, demonstrating that leveraging additional training resources for pretraining on disclosure documents consistently enhances model performance. CNAP consistently outperforms ROPT across all three document types, with further F1 score improvements of 0.3, 0.4, and 0.1 points on auditor's reports, IPO prospectuses, and annual reports, respectively. This performance gap can be attributed to the fact that tables, when processed separately following reading order, fail to provide sufficient supervision for identifying semantically equivalent numerical pairs.

\section{Conclusion and Future Work}
In this paper, we presented \textbf{CoFiTCheck}, a novel coarse-to-fine framework for document-level tabular numerical cross-checking with large language models. Our approach introduces two key technical innovations: (1) a contextualized instructional LLM encoder with parallel encoding and a decoupled InfoNCE objective for efficient, high-recall filtering of candidate pairs, and (2) a discriminative classification stage leveraging a specialized LLM with cross-table numerical alignment pretraining for fine-grained semantic matching. Extensive experiments on three types of real-world disclosure documents demonstrate that CoFiTCheck not only achieves state-of-the-art performance, but also delivers substantial improvements in efficiency, effectively addressing the core challenges of large-scale numerical semantic matching.


CoFiTCheck represents a significant advancement in automated tabular numerical cross-checking for disclosure documents, with promising implications for document-level fact verification in broader contexts. Future work could explore extending this framework to other table-rich domains that require accurate numerical consistency, such as scientific publications, regulatory filings, and technical reports. Since manual annotation is labor-intensive for numerical cross-checking tasks, it would be intriguing to investigate more sophisticated pretraining techniques to enhance model performance with limited labeled data.

\bibliographystyle{IEEEtran}
\bibliography{custom.bib}

\vspace{180pt}

\begin{IEEEbiography}[{\includegraphics[width=1in,height=1.25in,clip,keepaspectratio]{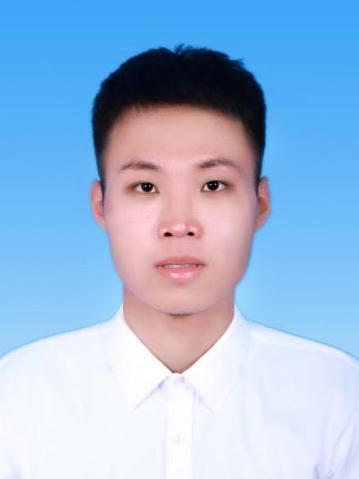}}]{Chaoxu Pang}
is currently a fourth-year PhD student at the Key Laboratory of Intelligent Information Processing, Institute of Computing Technology, Chinese Academy of Sciences (ICT, CAS), under the supervision of Professor Ping Luo. Before joining ICT, he received his bachelor's degree in information and communication engineering from Beijing University of Posts and Telecommunications. His research interests lie in natural language processing, large language models, and table understanding. He has published innovative works in top-tier conferences such as ACL and EMNLP.
\end{IEEEbiography}

\begin{IEEEbiography}[{\includegraphics[width=1in,height=1.25in,clip,keepaspectratio]{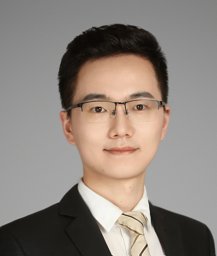}}]{Yixuan Cao}
is currently an associate professor in the Key Laboratory of Intelligent Information Processing, Institute of Computing Technology, Chinese Academy of Sciences (ICT, CAS). Before joining ICT, he received his Ph.D. in computer science from the University of Chinese Academy of Sciences (Institute of Computing Technology) in 2020. His research interests lie in natural language processing, document intelligence, and trustworthy AI. He has published papers in top-tier journals and conferences, such as KDD, WWW, CIKM, NeurIPS, and AAAI.
\end{IEEEbiography}

\begin{IEEEbiography}[{\includegraphics[width=1in,height=1.25in,clip,keepaspectratio]{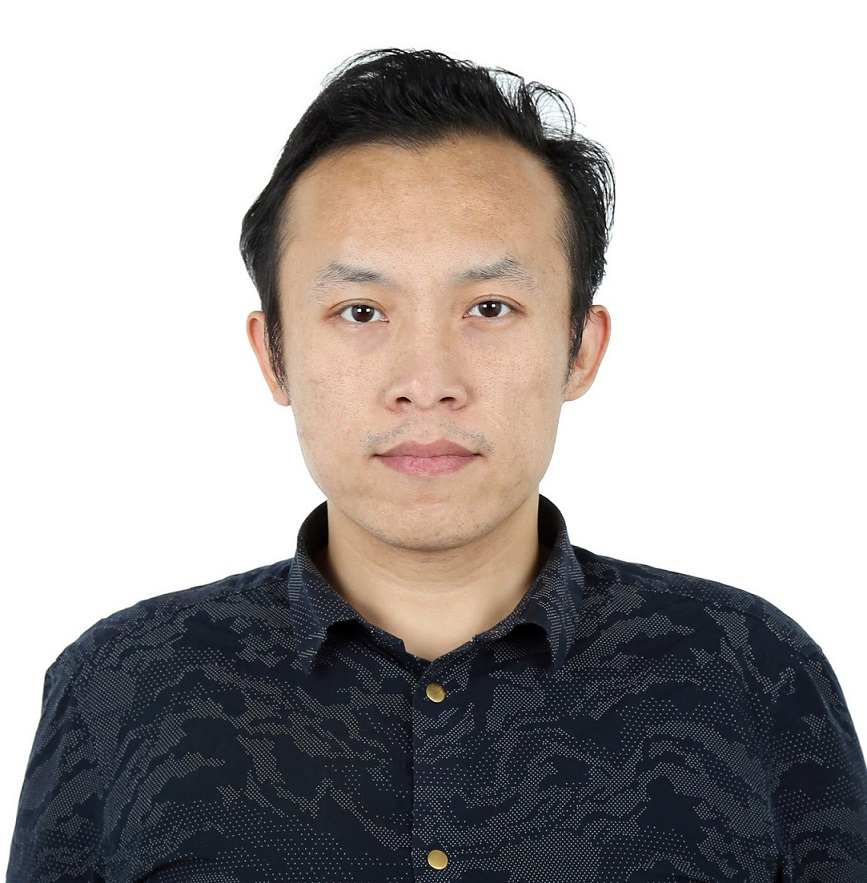}}]{Ganbin Zhou}
is currently an AI researcher at Beijing Paoding Technology Co., Ltd. (PAI Tech). He received his Ph.D. from the Institute of Computing Technology, Chinese Academy of Sciences in 2018. His main research interests include artificial intelligence, natural language processing, and information retrieval. He has published papers in top international conferences and journals such as AAAI and IJCAI.
\end{IEEEbiography}

\begin{IEEEbiography}[{\includegraphics[width=1in,height=1.25in,clip,keepaspectratio]{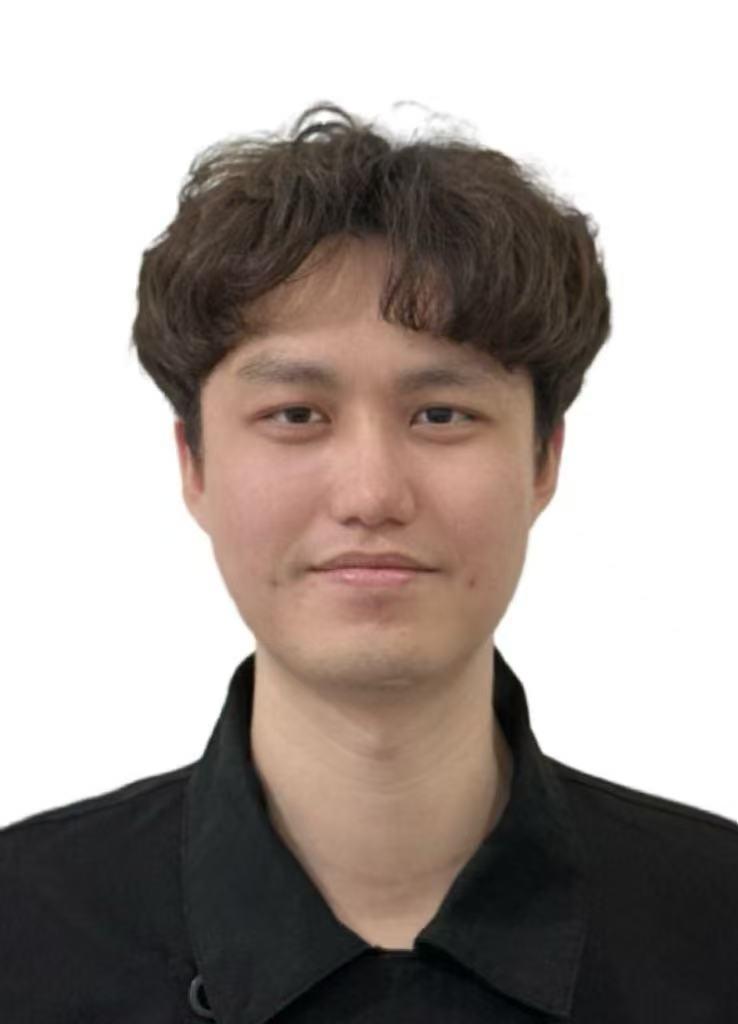}}]{Hongwei Li}
is currently an AI researcher at Beijing Paoding Technology Co., Ltd. (PAI Tech). He received his Ph.D. in computer science from the University of Chinese Academy of Sciences (Institute of Computing Technology) in 2020. His research interests include natural language processing, computer vision, and document intelligence. He has published papers in top international conferences and journals such as KDD and WWW.
\end{IEEEbiography}

\begin{IEEEbiography}[{\includegraphics[width=1in,height=1.25in,clip,keepaspectratio]{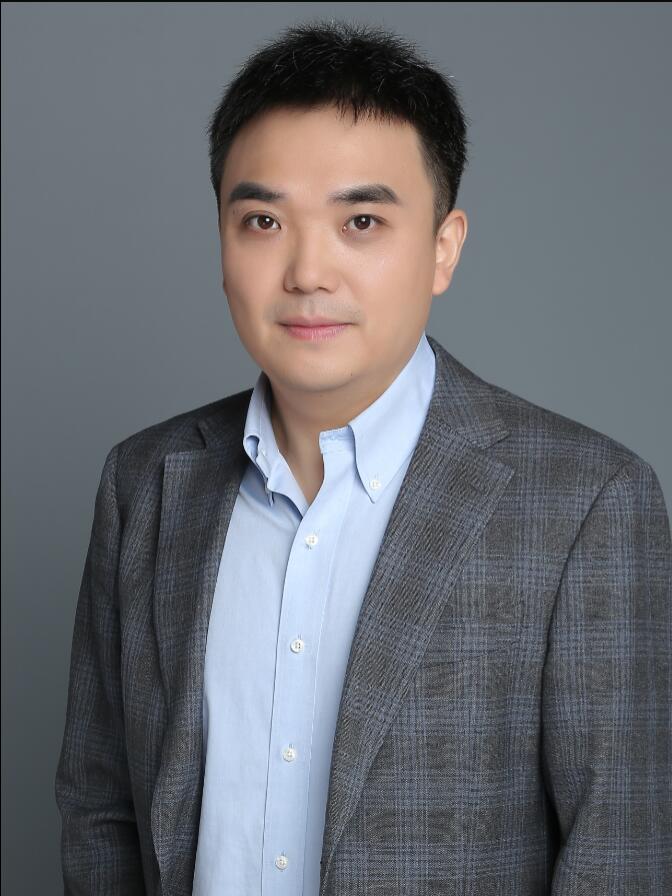}}]{Ping Luo}
is currently an associate professor at the Key Laboratory of Intelligent Information Processing, Institute of Computing Technology, Chinese Academy of Sciences (ICT, CAS), and the University of Chinese Academy of Sciences. Before joining ICT, he served as a senior research scientist and research manager at Hewlett-Packard Labs, China. His research interests include data mining and machine learning, with a particular focus on document AI. Dr. Luo has published over 100 research papers in top-tier journals and conferences, such as \emph{IEEE Transactions on Knowledge and Data Engineering}, \emph{IEEE Transactions on Information Theory}, KDD, CIKM, WSDM, ICDM, and NeurIPS. He has received several prestigious awards, including the ACM CIKM Best Student Paper Award (2012), ACM CIKM Best Paper Candidate Award (2010), SDM Best Paper Candidate Award (2010), and the Doctoral Dissertation Award from the China Computer Federation (2009).
\end{IEEEbiography}


\vfill

\end{document}